\documentclass[german,english]{bvm2019}

\title{Enhancing Label-Driven Deep Deformable Image Registration with Local Distance Metrics for State-of-the-Art Cardiac Motion Tracking}

\titlerunning{Enhancing Label-Driven Deep Deformable Image Registration}

\author{Alessa~Hering$^1$$^,$$^2$, Sven~Kuckertz$^1$$^,$$^3$, Stefan Heldmann$^1 $, Mattias~P.~Heinrich$^3$}
\authorrunning{Hering et al.}

\institute{%
$^1$Fraunhofer MEVIS, L\"ubeck\\
$^2$Diagnostic Image Analysis Group, Radboud UMC, Nijmegen, Netherlands\\
$^3$Institute of Medical Informatics, University of L\"ubeck}

\email{alessa.hering@mevis.fraunhofer.de}

\newcommand{\F}{\ensuremath{\mathcal{F}}}
\newcommand{\M}{\ensuremath{\mathcal{M}}}

\usepackage{booktabs}
\usepackage{xcolor}
\usepackage{colortbl}
\usepackage{amsmath}

\definecolor{LVCcolor}{RGB}{0,90,91}
\definecolor{RVCcolor}{RGB}{181,22,33}
\definecolor{LVMcolor}{RGB}{250,187,0}
\definecolor{Meancolor}{RGB}{144,99,41}

\definecolor{colE1}{RGB}{129,188,14}
\definecolor{colE2}{RGB}{187,222,7}
\definecolor{colE3}{RGB}{210,211,0}
\definecolor{colE4}{RGB}{243,208,0}
\definecolor{colE5}{RGB}{250,187,0}
\definecolor{colE6}{RGB}{238,145,8}
\definecolor{colE7}{RGB}{213,87,19}
\definecolor{colE8}{RGB}{181,22,33}

\begin{document}

%
\selectlanguage{english}

\maketitle

\begin{abstract}
While deep learning has achieved significant advances in accuracy for medical image segmentation, its benefits for deformable image registration have so far remained limited to reduced computation times. Previous work has either focused on replacing the iterative optimization of distance and smoothness terms with CNN-layers or using supervised approaches driven by labels. Our method is the first to combine the complementary strengths of global semantic information (represented by segmentation labels) and local distance metrics that help align surrounding structures. We demonstrate significant higher Dice scores (of 86.5\%) for deformable cardiac image registration compared to classic registration (79.0\%) as well as label-driven deep learning frameworks (83.4\%).

\end{abstract}

\section{Introduction}
Image registration aims to align two or more images to achieve point-wise spatial correspondence. This is a fundamental step for many medical image analysis tasks and has been a very active field of research for decades. In deformable image registration approaches, non-rigid, non-linear deformation fields are established between a pair of images, such as cardiac cine-MR images. Typically, image registration is phrased as an unsupervised optimization problem w.r.t. a spatial mapping that minimizes a suitable cost function by applying iterative optimization schemes. Due to substantially increased computational power and availability of image data over the last years, learning-based image registration methods have emerged as an alternative to energy-optimization approaches.

\textbf{Prior work on CNN-Based Deformable Registration:}
Compared to other fields relatively little research has yet been undertaken in deep-learning-based image registration. Only recently have the first deep-learning based image registration methods been proposed \cite{rohe2017svf,de2017DIRnet,hu2018weakly}, which mostly aim to learn a function in form of a CNN that predicts a spatial deformation warping a moving image to a fixed image. We categorize these approaches into \textit{supervised} \cite{rohe2017svf}, \textit{unsupervised} \cite{de2017DIRnet,krebs2018unsupervised} and \textit{weakly-supervised} \cite{hu2018weakly} techniques based on how they train the network.

The \textit{supervised} methods use ground-truth deformation fields for training. These deformation fields can either be randomly generated or produced by classic image registration methods. The main limitation of these approaches is that their performance is limited by the performance of existing algorithms or simulations. In contrast, the \textit{unsupervised} method do not require any ground-truth data. The learning process is driven by image similarity measures or more general by evaluating the cost function of classic variational image registration methods. An important milestone for the development of these methods was the introduction of the spatial transformer networks \cite{jaderberg2015spatial} to differentiably warp the moving image during training. \textit{Weakly-supervised} methods also do not rely on ground-truth deformation fields but training is still supervised with prior information. The labels of the moving image are transformed by the deformation field and compared within the loss function with the fixed labels. All anatomical labels are only required during training.

\textbf{Contributions:} 
We propose a new deep-learning-based image registration method that learns a registration function in form of a CNN to predict a spatial deformation that warps a moving image to a fixed image. In contrast to previous work, we propose the first weakly-supervised approach, which successfully combines the strengths of prior information (segmentation labels) with an energy-based distance metric within a comprehensive multi-level deep-learning based registration approach. 
\section{Materials and Methods}
\begin{figure}
\centering
\includegraphics[width=0.8\textwidth]{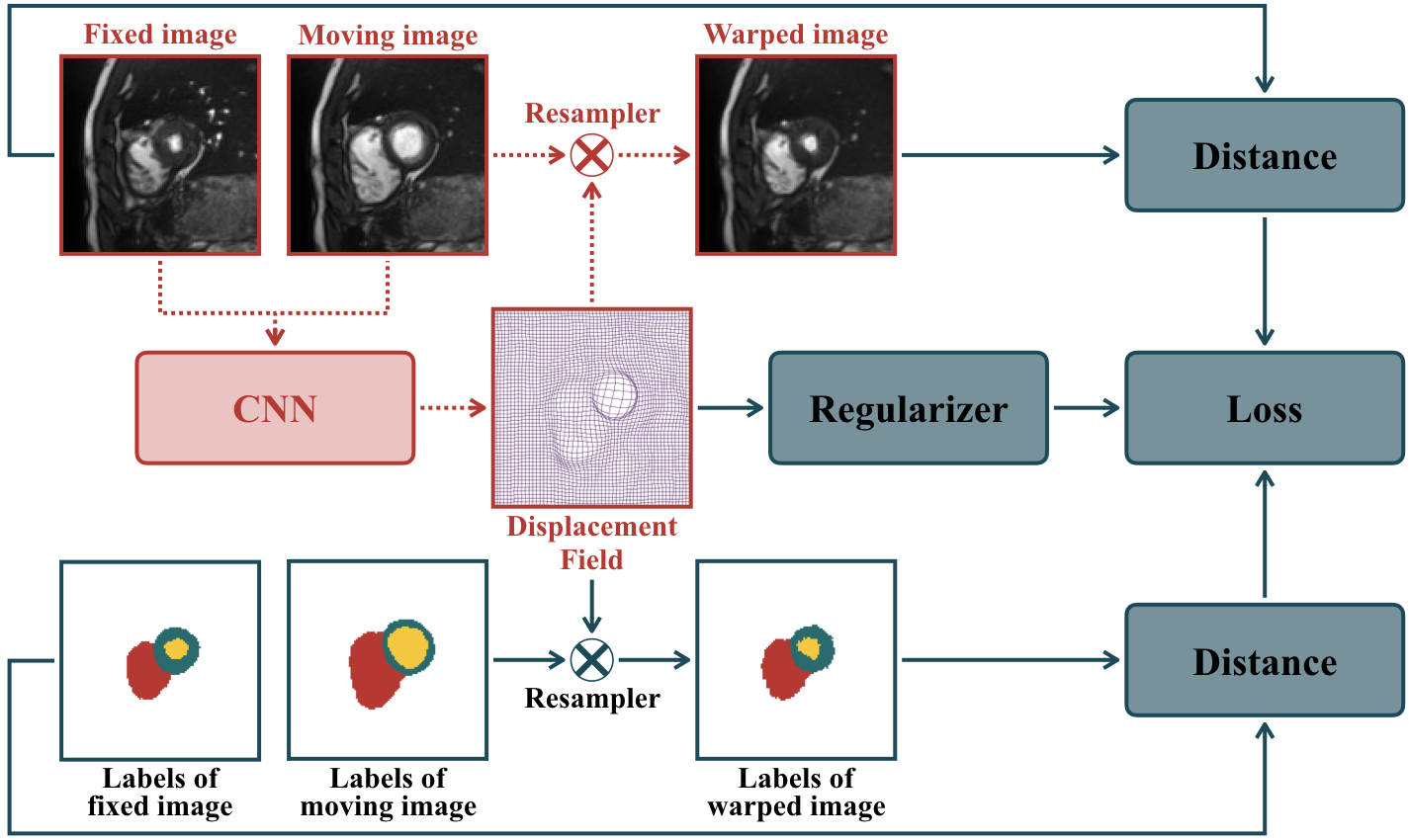}
\caption{Illustration of the training process. For convenience, there is only one output deformation field shown instead of three. While application after the training only flows represented by red-dotted lines and red parts are required.}
\label{fig:Aufbau}
\end{figure}
Let $\mathcal{F},\mathcal{M}:\mathbb{R}^2\to\mathbb{R}$ denote the fixed image and moving image, respectively, and let $\Omega\subset\mathbb{R}^2$ be a domain modeling the field of view of $\mathcal{F}$. We aim to compute a deformation $y:\Omega\to\mathbb{R}^2$ that aligns the fixed image $\mathcal{F}$ and the moving image $\mathcal{M}$ on the field of view $\Omega$ such that $\mathcal{F}(x)$ and $\mathcal{M}(y(x))$ are similar for $x\in\Omega$. Inspired by recent unsupervised image registration methods (e.g. \cite{de2017DIRnet,rohe2017svf}), we do not employ iterative optimization like in classic registration, but rather use a CNN that takes images $\F$ and $\M$ as input and yields the deformation $y$ as output (cf. Fig.~\ref{fig:Aufbau}). Thus, in the context of CNNs, we can consider $y$ as a function of input images \F,\M and trainable CNN model parameters $\theta$ to be learned, i.e. $
    y(x) \equiv y(\theta;\F,\M,x).$
During the training, the CNN parameters $\theta$ are learned so that the deformation field $y$ minimizes the loss function
\begin{equation}
\mathcal{L}(\F,\M,b_\F,b_\M,y)~=~\delta\cdot\mathcal{D}(\F,\M(y))~+~\alpha\cdot\mathcal{R}(y)~+\beta\cdot\mathcal{B}(b_\F,b_\M(y))
\label{eq:loss}
\end{equation} with so-called distance measure $\mathcal{D}$ that quantifies the similarity of fixed image $\F$ and deformed moving image $\M(y)$, regularizer $\mathcal{R}$ that forces smoothness of the deformation and a second distance measure $\mathcal{B}$ that quantifies the similarity of fixed segmentation $b_\F$ and warped moving segmentation $b_\M(y)$. The parameters $\delta, \alpha, \beta\ge0$ are weighting factors. For convenience, we set $\delta=1$. Note that the segmentations are only used to evaluate the loss function and not used as network input and are therefore only required during training. 
We use the edge-based normalized gradient fields distance measure  \cite{ruhaak2013highly}
\begin{equation*}
    \mathcal{D}(\F,\M(y)) = \frac{1}{2}\int_\Omega 1- \frac{\langle \nabla \M(y(x)), \nabla \F(x)\rangle^2_\epsilon}{\Vert\nabla\M(y(x))\Vert^2_\epsilon \Vert\nabla\F(x)\Vert^2_\epsilon} \, \text{d}x,
\end{equation*}
with $\langle f,g \rangle_\epsilon := \sum_{j=1}^2 f_j g_j +\epsilon^2$, 
$\|f\|_\epsilon := \sqrt{\langle f, f \rangle_\epsilon}$, so-called edge parameter $\epsilon>0$ and curvature regularizer $
    \mathcal{R}(y) = \frac{1}{2} \int_\Omega \sum_{j=1}^2 \Vert \Delta y_j \Vert^2 \, \text{d}x $ \cite{ruhaak2013highly}.
The similarity of the segmentation masks is measured using a sum of squared differences of the one-hot-representation of the segmentations 
$\mathcal{B}(y) =  \frac{1}{2} \int_\Omega \| b_\M(y(x)) - b_\F(x)) \|^2 \text{d}x.$ 

\begin{figure}
\centering
\includegraphics[width=0.9\textwidth]{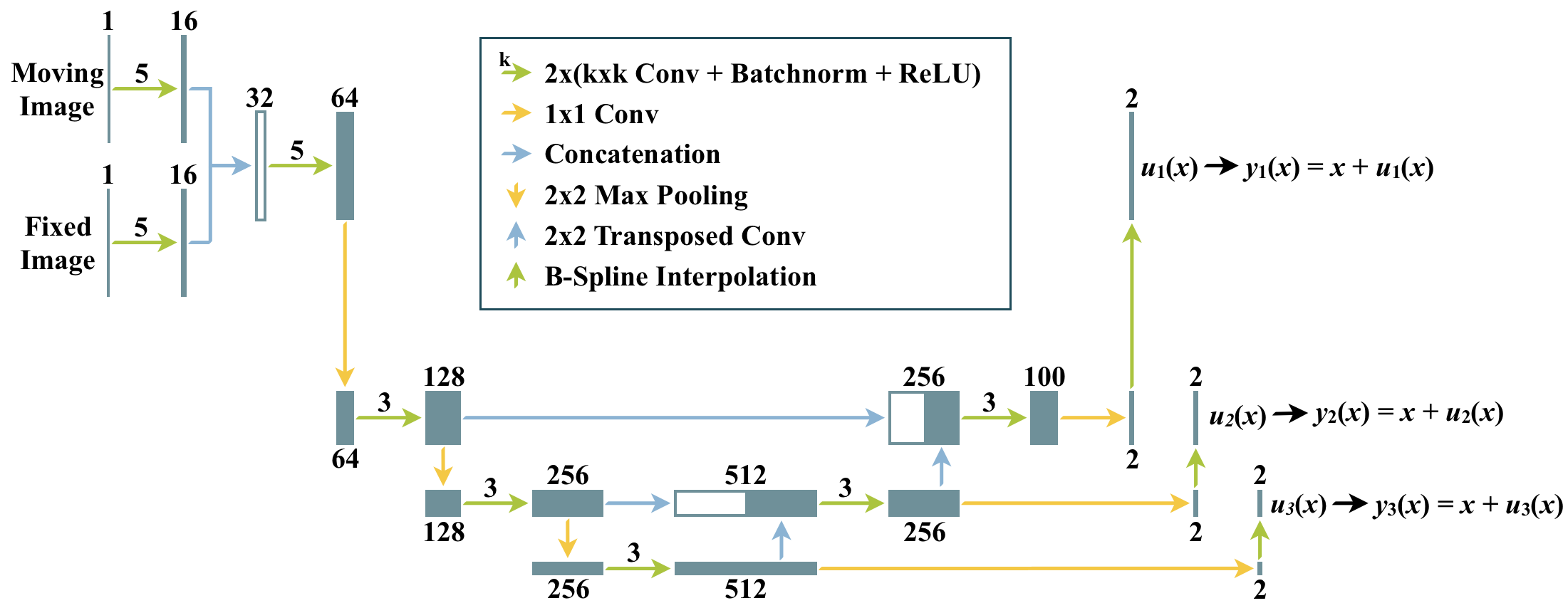}
\caption{Proposed UNet based architecture of our CNN. Each blue box represents a multi-channel feature map whose width corresponds corresponds to the number channels which is denoted above or below the box.}
\label{fig:UNet}
\end{figure}%
\begin{figure}
\centering
\includegraphics[width=0.75\textwidth]{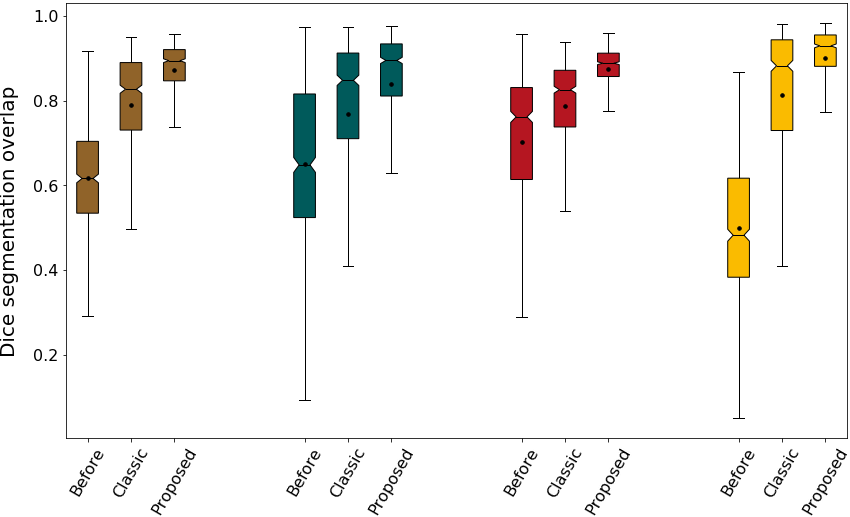}
\caption{Comparison of Dice overlaps for all test images and each anatomical label (\textcolor{Meancolor}{\rule{.2cm}{.2cm}} average of all labels, \textcolor{LVCcolor}{\rule{.2cm}{.2cm}} left ventricle cavity (vc), \textcolor{RVCcolor}{\rule{.2cm}{.2cm}} right vc and \textcolor{LVMcolor}{\rule{.2cm}{.2cm}} myocardium). For each one the distributions of Dice coefficients before, after classic and after our proposed registration are shown.}
\label{fig:boxplots}
\end{figure}%
\textbf{Architecture and Training:} Our architecture is illustrated in Fig.~\ref{fig:UNet}. Our network architecture basically follows the structure of a UNet \cite{ronneberger2015unet}, taking a pair of fixed and moving images as input. However, we start with two separate, yet shared processing streams for the moving and fixed image. The CNN generates a grid of control points for a B-spline transformer, which output is a full displacement field to warp a moving image to a fixed image. During training, the outputs of the network are three deformation fields of different resolutions. We compute the overall loss as a weighted sum of the network outputs on this different resolution levels. This design decision is inspired by the multi-level strategy of classic image registration. During inference, only the deformation field on the highest resolution is used.

\textbf{Experiments:} We perform our experiments on the ACDC dataset \cite{ACDC2017}. It contains cardiac multi-slice 2D cine-MR images of 100 patients captured at end-diastole (ED) and end-systole (ES) time point, amounting to 951 2D images per cardiac phase. The dataset includes annotations for left and right ventricle cavity and myocardium of a clinical expert. We only use slices that contain all labels, i.e. 680 2D images pairs. All images are cropped to the region of interest with a size of 112x112 pixels. Image intensities are normalized to a range of $\left[ 0,1 \right]$. For data augmentation we slightly deform the images and segmentations to increase the number of image pairs by a factor of 8. 

Training is performed as a $k$-fold cross-validation with $k=10$ which divide our dataset patient-wise. Our method is implemented in PyTorch. The network is trained for 40 epochs on a NVIDIA GTX 1080 in approximately 0.5 hours using an ADAM optimizer with a learning rate of $10^{-3}$ and a batch size of 30. We empirically choose the regularization parameter $\alpha= 10^3$, the boundary-loss weight $\beta= 5\cdot10^4$ and edge parameter $\epsilon= 0.1$ in the loss function. To evaluate our registration method we use the computed deformation field to propagate the segmentation of the moving image and measure the volume overlap using the Dice coefficient. We compare our method with a classic multi-level image registration model similar to \cite{ruhaak2013highly} which iteratively minimizes the loss function without the use of segmentation data.

\section{Results}
\begin{table}
\caption{The quantitative effect of variations of the weighting within the loss function $\mathcal{L}~=~\delta\cdot\mathcal{D}~+~\alpha\cdot\mathcal{R}~+~\beta\cdot\mathcal{B}$ is shown when varying one parameter and fixing the others to their empirically determined optimal values. Besides the resulting Dice coefficient the percentage of pixels in which foldings ($\text{det}(\nabla y)\leq 0$) occur is depicted. \label{tab:variation}}
\begin{minipage}{.85\linewidth}
\centering
\begin{tabular*}{\linewidth}{l @{\extracolsep{\fill}} cccccc}
\toprule 
opt. $\rightarrow$& \multicolumn{2}{c}{$\delta=1$} & \multicolumn{2}{c}{$\alpha=10^3$} & \multicolumn{2}{c}{$\beta=5\cdot10^4$} \\ \midrule 
& Dice & Folding & Dice & Folding & Dice & Folding\\
\midrule

$\times10^{2}$  & \cellcolor{colE8!85}0.1\%     & 78.2\%    & \cellcolor{colE8!85}75.4\%    & 0.0\%     & \cellcolor{colE2!85}86.0\%    & 0.6\% \\
$\times10^{1}$  & \cellcolor{colE5!85}83.1\%    & 0.7\%     & \cellcolor{colE5!85}83.5\%    & 0.2\%     & \cellcolor{colE1!85}87.2\%    & 0.1\% \\ \midrule
                & \cellcolor{colE2!85}86.5\%    & 0.0\%     & \cellcolor{colE2!85}86.5\%    & 0.0\%     & \cellcolor{colE2!85}86.5\%    & 0.0\% \\ \midrule
$\times10^{-1}$ & \cellcolor{colE3!85}85.9\%    & 0.0\%     & \cellcolor{colE1!85}87.0\%    & 0.2\%     & \cellcolor{colE3!85}85.3\%    & 0.1\% \\
$\times10^{-2}$ & \cellcolor{colE3!85}85.5\%    & 0.1\%     & \cellcolor{colE1!85}87.2\%    & 0.4\%     & \cellcolor{colE7!85}81.6\%    & 0.4\% \\
$\times 0$      & \cellcolor{colE5!85}83.4\%    & 0.1\%     & \cellcolor{colE8!85}0.0\%     & 40.6\%    & \cellcolor{colE7!85}81.9\%    & 0.4\% \\\bottomrule

\end{tabular*}
\end{minipage}%
\begin{minipage}{.15\linewidth}
  \centering
    \begin{tabular}{c}
    Colormap\\ \toprule
    \cellcolor{colE1!85}87\%\\ 
    \cellcolor{colE2!85}86\%\\
    \cellcolor{colE3!85}85\%\\
    \cellcolor{colE4!85}84\%\\
    \cellcolor{colE5!85}83\%\\ 
    \cellcolor{colE6!85}82\%\\ 
    \cellcolor{colE7!85}81\%\\
    \cellcolor{colE8!85}$\le$80\%\\
    \end{tabular}
\end{minipage} 
\end{table}

\begin{figure}[t]
\centering
\begin{tabular}{ccccc}
  \includegraphics[width=0.17\textwidth]{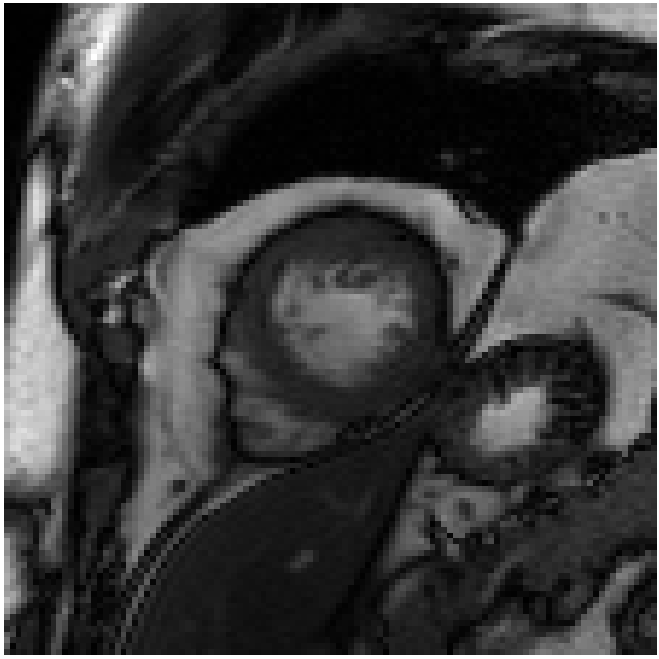}
& \includegraphics[width=0.17\textwidth]{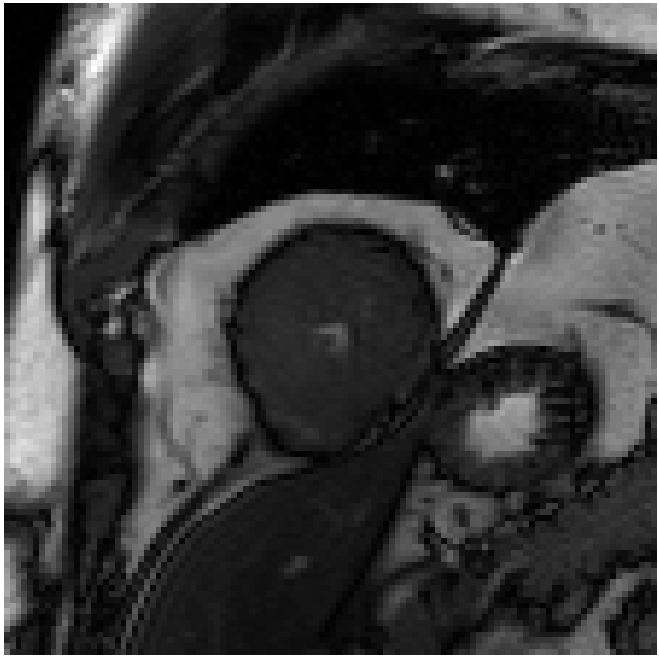}
& \includegraphics[width=0.17\textwidth]{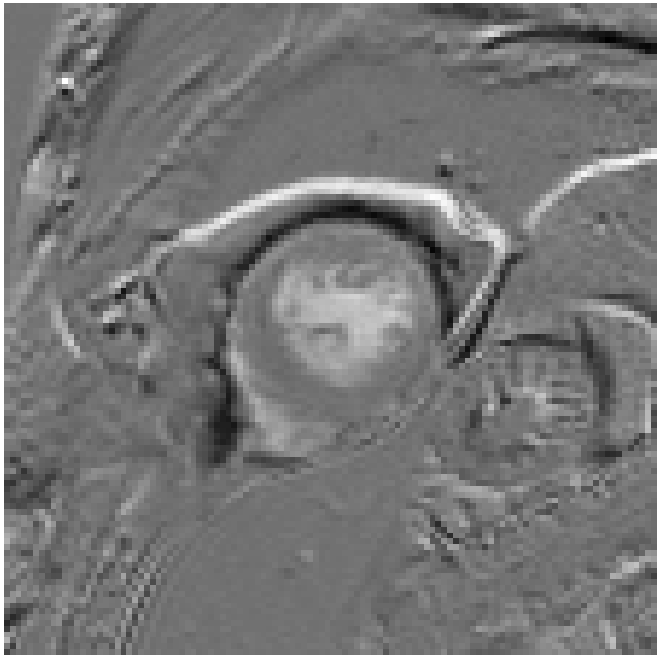}
& \includegraphics[width=0.17\textwidth]{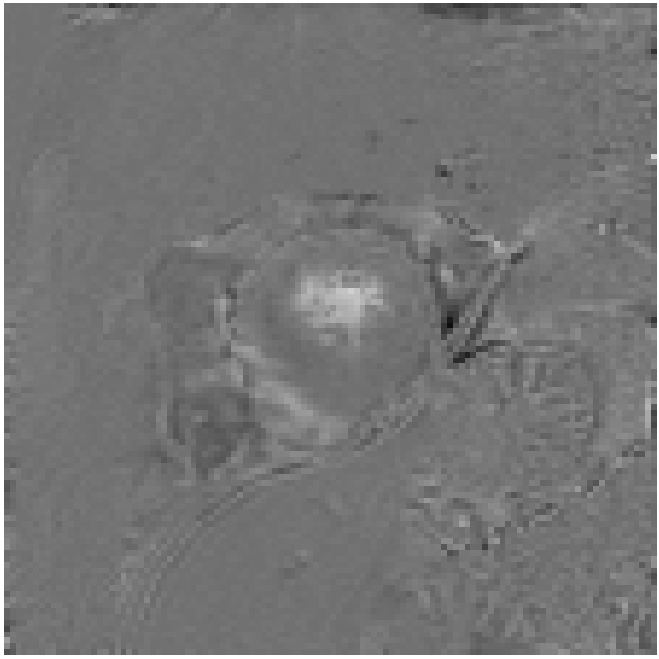}
& \includegraphics[width=0.17\textwidth]{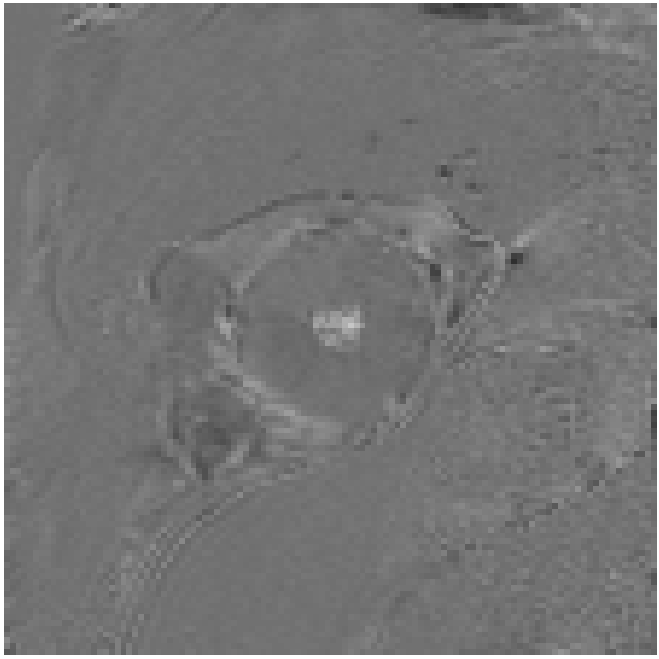}
\\
  \includegraphics[width=0.17\textwidth]{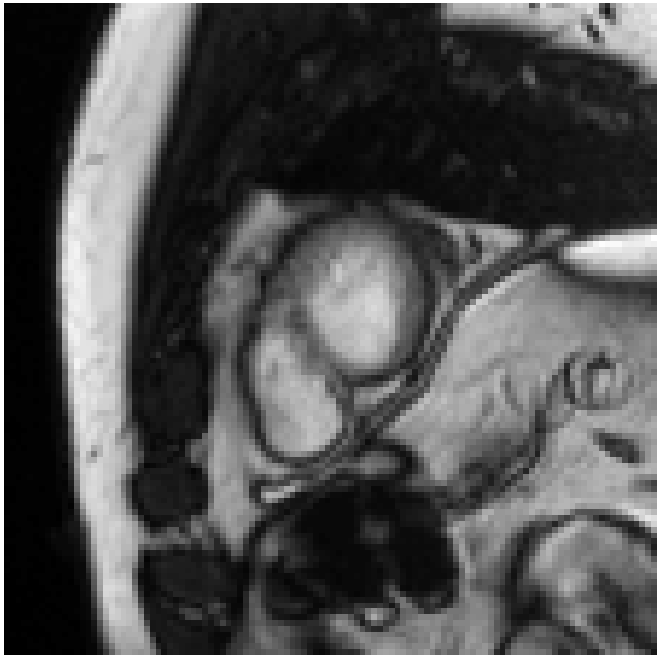}
& \includegraphics[width=0.17\textwidth]{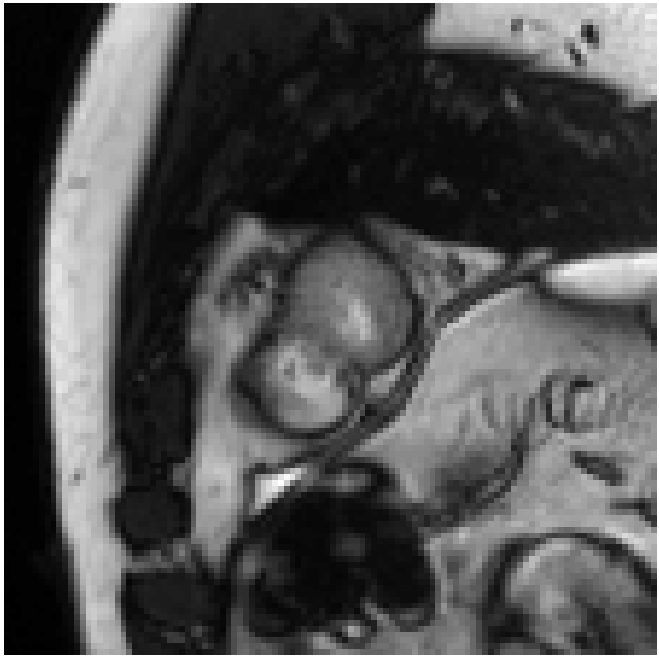}
& \includegraphics[width=0.17\textwidth]{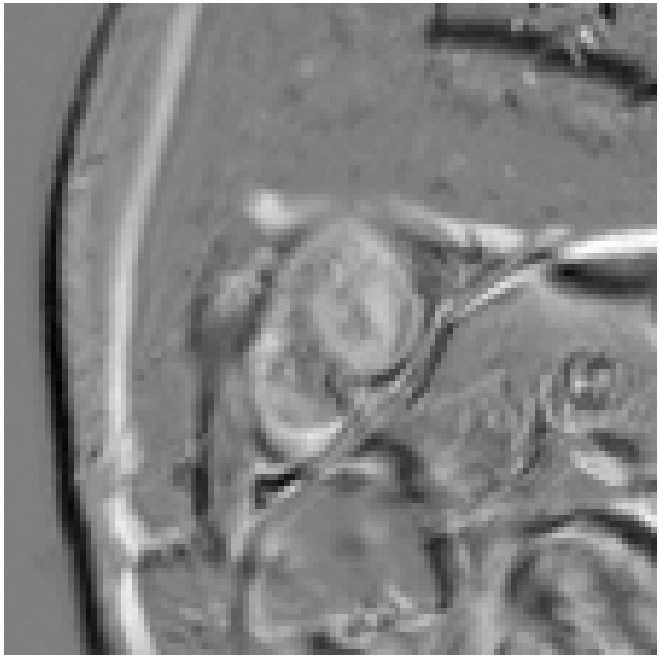}
& \includegraphics[width=0.17\textwidth]{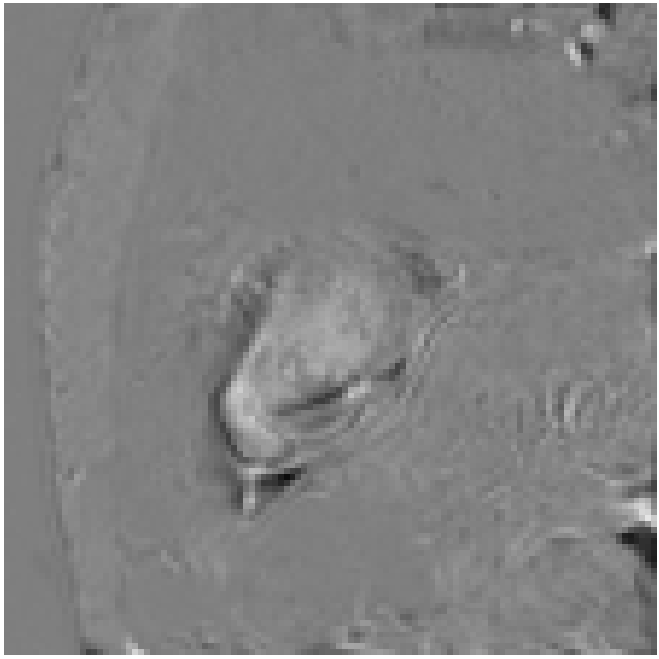}
& \includegraphics[width=0.17\textwidth]{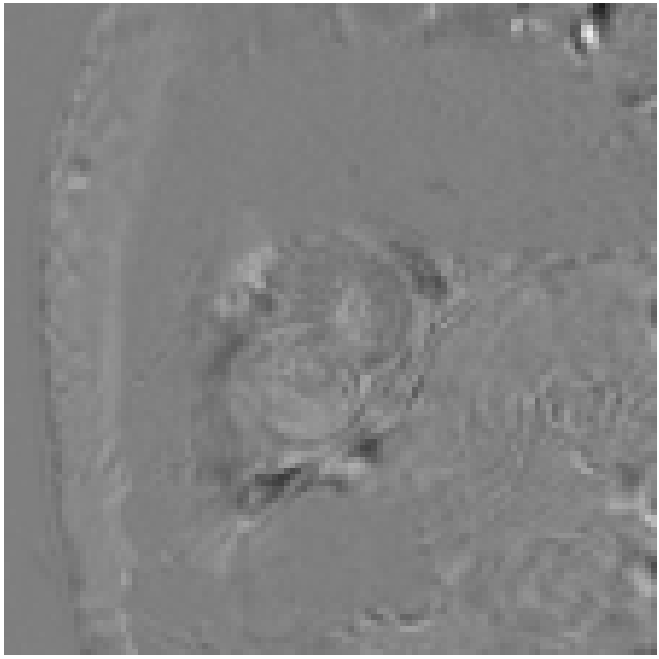}
\\
  {Moving $\M$}
& {Fixed $\F$}
& {$\M-\F$}
& {$\M(y_\text{classic})-\F$}
& {$\M(y_\text{prop})-\F$}
\end{tabular}
\caption{Example input images $\M$ and $\F$, difference image of the input images (third column), of fixed image and the warped image after classic registration (fourth column) and of fixed image and warped image after our proposed registration (fifth column). White and black indicate a great difference, while grey means similar images.} \label{fig:results}
\end{figure}

As shown in Fig.~\ref{fig:boxplots}, our proposed method outperforms the classic multi-level image registration approach compared by the average Dice coefficient. Our method achieves an average improvement from $79.0\%$ to $86.5\%$ across all three labels in terms of Dice accuracy while reducing the computation time drastically from 3.583s to 0.006s per image pair. 
Not only the average Dice score of our approach is higher for every anatomical label, but also the variation is reduced (see Fig.~\ref{fig:boxplots}). Fig~\ref{fig:results} shows two example image pairs and the registration results, demonstrating the ability of our method to compensate large local deformations.\\
In comparison to the method of Krebs \cite{krebs2018unsupervised} which uses the same dataset and present comparable Dice coefficients of unregistered images, our method yields an improvement from $78.3\%$ to $86.5\%$.
As illustrated in Table~\ref{tab:variation}, our choice of weighting parameters within the loss function~\eqref{eq:loss} leads to a compromise between maximizing the Dice score and keeping the percentage of foldings low. 

\section{Discussion}

We have presented a new weakly-supervised deep-learning-based method for image registration that replaces iterative optimization steps with deep CNN-layers. Our approach advances the state-of-the-art in CNN-based deformable registration by combining the complementary strengths of global semantic information (supervised learning with segmentation labels) and local distance metrics borrowed from classical medical image registration that supports the alignment of surrounding structures. We have evaluated our technique on dozens of cardiac multi-slice 2D cine-MR images and demonstrate substantial improvements compared to classic image registration methods both in terms of Dice coefficient and execution time. We also demonstrated the importance of integrating both unsupervised (distance measure $\mathcal{D}$) supervised (boundary term $\mathcal{B}$) learning objectives into a unified framework and achieve state-of-the-art Dice scores of 86.5\%.

\bibliographystyle{bvm2019}

\bibliography{2683}
\end{document}